\documentclass[preprint,12pt]{elsarticle}
\usepackage{subfig}
\usepackage[margin=1in]{geometry} 
\usepackage{subfloat}
\usepackage{tabularx,ragged2e}
\newcolumntype{Y}{>{\RaggedRight\arraybackslash\hspace{0pt}}X}
\usepackage{amsmath}

\usepackage{float}
\usepackage{amssymb}

\journal{Journal Name}

\begin{document}

\begin{frontmatter}


\title{A data-driven personalized smart lighting recommender system }

\author{Atousa Zarindast\corref{cor1}\fnref{label1}}


\author[label1]{Jonathan Wood}
\author[label1]{Anuj Sharma}

\address[label1]{Department of civil and environment egineering, Iowa state university, Ames, Iowa, USA }



\begin{abstract}
Recommender systems attempts to identify and recommend the most preferable item (product-service) to an individual user. These systems predict  user interest in items based on related items,  users, and the interactions between items and users \cite{lu2015recommender}. We aim to build an auto-routine and color scheme recommender system that leverages a wealth of historical data and machine learning methods. We introduce an unsupervised method to recommend a routine for lighting. Moreover, by analyzing users' daily logs, geographical location, temporal and usage information we understand user preference and predict their preferred color for lights. To do so, we cluster users based on their geographical information and usage distribution. We then build and train a predictive model within each cluster and aggregate the results. Results indicate that models based on similar users increases the prediction accuracy, with and without prior knowledge about user preferences.

\end{abstract}


\end{frontmatter}


\section{Introduction}
\label{S:1}
Recommender systems have many application domains including advertisement, e-commerce, etc. Outlining the desired outcomes and objectives is an essential step in developing quality recommender systems. For instance, in traditional e-commerce recommender systems, the desired outcome is typically met by suggesting items that customers are likely to purchase. On the other hand, the customers' goals are somehow indirectly in alignment with the recommender system and, as a result, an increase in direct sales would increase organization profits. Designing an effective recommender system is a challenging problem due to data input quality issues and prioritization of desired outcomes (e.g., prediction accuracy, novelty of product, sales increase, etc.). Achieving high accuracy metrics is ideal, but the quality of the recommendations is influenced by additional factors such as confidence, novelty, risk, privacy, reliability, etc \cite{shani2011evaluating}. Hence, we first need to define recommender system goals in collaboration with stakeholders' interests so that we can prioritize the focus in our framework.

The infrastructure we currently have in place allows us to learn from past usage data and propose personalized recommendations in smart lighting business. A network of connected things - known as the Internet of things - IoT is filling the gap between lighting and connected grid systems via smart lighting systems. Smart lighting is a promising technology for home energy management, mood improvement \cite{boyce2000mood}, and improved sleep quality by enabling brightness adjustment  (e.g., brightly lit homes in the evening have been shown to adversely impact circadian rhythms and sleep) \cite{cain2020sleep}. Smart lighting and control systems include information and communication technologies such as wireless networks, sensors and light-emitting diodes (LEDs), and link with other smart infrastructure devices (e.g., Google Nest)  \cite{pandharipande2018lighting}. As mentioned earlier, a potential benefit of smart lighting systems is energy savings. Previous studies have focused on energy saving through light-emitting diodes (LED) \cite{peruffo2015lighting,zou2018winlight}. These smart lighting systems, as compared to traditional lighting systems, are installed in office locations and have high potential for energy savings (i.e., 17-60\% depending on usage patterns) \cite{von2001analysis}. Machine learning has shown a favorable impact in different disciplines such as traffic engineering \cite{zarindast2021big,atousa:online} , health care \cite{rajabalizadeh2020exploratory}, and sociology \cite{moeinizade2019predicting}. With recent developments in artificial intelligence and advancement in computational power and storage capability, it is now possible to leverage historical data, learn patterns and provide personalized recommendations.

Public interest in smart lighting systems is growing. Studies have revealed that working under desired lighting should increase satisfaction and productivity levels \cite{juslen2007influence}. Moreover, suitable visual lighting and color can reduce stress levels and affect mood \cite{dalke2006colour}. In fact, color temperature has been proven to have a huge impact on human psychological processes \cite{knez2001effects}. Therefore, providing personalized lighting recommendations could increase user satisfaction and user engagement with the lighting system. Every individual's preferences, needs, and tolerance levels in terms of lighting and color is different \cite{tregenza1974consistency}. Even for a particular user, different tasks may require different lighting settings. For example, high color temperature could result in higher alertness and increased productivity \cite{park2010study} due to a higher proportion of blue light spectrum (460–500 nm) which prevents melatonin secretion.
\cite{webb2006considerations}. In that way, higher alertness is suitable for working conditions, but it is not suitable for sleeping. Ideally, lighting recommendations should consider not only environmental factors such as daylight \cite{pazzani2007content} but also individual lighting preferences based on visual and emotional needs and inclinations \cite{bellia2011lighting}.

In this paper we aim to include recommendation systems as a smart add-in feature in products in order to increase customer utility, loyalty and satisfaction as they interact with the system. Added advantages, such as routine and scene recommendations, can result in increased interest in smart lighting, which would increase the adoption rate. Consequently, increases in adoption rates would reduce the production costs and help increase annual net profits. A reduction in production costs defined by economic scale can create a business opportunity for integrated sensors. The implementation of such sensors, along with lighting routines and schedules defined in smart lighting systems can help with energy saving. Thus, by recommending a lighting routine and introducing a smart enterprise lighting system, we aim to contribute to energy savings. In order to reflect user preferences, we leverage historical data to understand user behavior. We utilize terabytes of daily user logs while interacting with the system to analyze user behavior, character, and preferences to obtain contextual information. Finally, using machine learning, we build a context-aware predictive model based on geographical, temporal, and usage data of lights in households. The end product of this system is a light color recommendation, defined as "scene". By sampling a diverse and wealthy amount of data in terms of time period and number of households, our solution is robust as its metrics show low bias and high confidence levels.

\section{Literature review}
\label{S:3}
\subsection{Light control systems}
Quality lighting can contribute to mood and productivity while maintaining comfort and increasing satisfaction \cite{juslen2007influence,van2014critical,boyce2000mood}. Conversely, inappropriate lighting  can affect  productivity, health\cite{bellia2011lighting}, satisfaction \cite{rea2000iesna}, and sleep quality \cite{cain2020sleep}.  A significant amount of people's time is spent indoors and hence one of the main objectives of building control systems should be to provide indoor comfort. However, building control systems usually neglect occupant satisfaction in lighting design criteria \cite{park2018comprehensive}. One aspect of comfort is defined as having control over indoor environment while interacting with the system \cite{nagy2016occupant}.

Occupant based control systems are a feasible solution for current automated control systems in work places. There are several occupancy based light control system studies specific for work places. For example, \citet{caicedo2011occupancy} suggested a light control system that considers occupancy and location of occupant to provide optimum brightness level. \citet{peruffo2015lighting} proposed a wireless network lighting system with multiple sensors that determines daylight and occupancy along with a central controller. The output of the control system is a dimming level of lights that is based on occupancy and daylight. \citet{gunay2017development} proposed a lighting and blind control algorithm for office environment. Moreover, they analyzed occupant behavior with different scenarios and simulations. These models automatically control lighting systems but neglect human perception, mood, comfort and preferences. Given the importance of user satisfaction and comfort,  research on both comfort and control systems has emerged \cite{heydarian2015immersive,de2015personal}. 

Recent studies have conducted research on lighting control systems based on occupancy, and have considered occupant comfort in their modeling for office layouts.
\citet{nagy2015occupant} introduced an occupancy based lighting control system based on statistical analysis and sensor data. They identified minimum and maximum thresholds brightness levels by interacting with the user and with statistical data analysis.
\citet{cheng2016satisfaction} proposed a closed loop satisfaction based Q-learning control system that receives users perception as feedback signals. In the proposed Q-learning based system, users' explicit feedback and interaction with the system is required. User feedback is always a valid source of modeling, but is not always practical to incorporate into system. Users may not be willing to take the time to give feedback for the system particularly in home based environment. Therefore, implicit understanding of user preferences while interacting with the system is likely a more practical and feasible solution is residential settings.

\citet{zou2018winlight} proposed a smart lighting control system that adjusts the brightness level of lamps based on real time occupancy data to minimize energy consumption. They included personalization by an app control feature that enables the occupant to adjust the brightness level of a nearby lamp. They did an experiment on a 24-week time period and evaluated the performance based on occupancy detection accuracy and energy saving. In their proposed model, the desired brightness level of each occupant is a given parameter to an optimization problem and it does not reflect the dynamics in mood, preferences and different requirements for different activities.
\citet{park2019lightlearn} introduced a reinforcement learning based control system for office location that is based on occupancy. They collected data from 5 offices for 8 weeks.  All of the above mentioned studies propose a control system for business framework which would not be suitable for household requirements, but are useful to understand.

\subsection{Recommendation systems}
Recommendation systems can be defined as attempts to identify and recommend the most preferable item (product-service) to an individual user. Commonly used recommendation techniques include collaborative filtering \cite{burke2000knowledge} which suffers from sparseness, scalability and cold start problems \cite{adomavicius2005toward}, content' based \cite{pazzani2007content} techniques which have overspecialized recommendations. Content based filtering primarily extracts the content as a basis for item prediction and attempts to build a user profile using preference indicators \cite{van2019recommender}.There is variety of applications for recommender systems including E-library \cite{porcel2010dealing}, E-commerce \cite{lawrence2001personalization}, movie, video \cite{lee2010collaborative} and TV program \cite{kwon2011personalized} recommenders. To this date, however, no studies have provided recommendation systems
or investigated machine learning to recommend light usage routine and scenes. 

One of the challenges in recommendation systems is the cold-start problem and which is due to lack of prior knowledge specific to an individual new user. To deal with cold-start, researchers have recently considered social media as a source of understanding customer characteristics and traits. Recently \citet{cho2020platform} included what is called lifelog information in smart lighting control system. Lifelogs include personal information related to a user's activities, biometrics and environmental information. They considered a user's message data, app location and activity data plus weather data in their analysis and introduced a lighting control system. They set up their system for one particular household using an infrastructure consisting of  motion sensor, pressure sensor on seats, an IoT camera in the kitchen for taking pictures of the food consumed by the user. Moverover, to understand the user's emotions, they utilized text message analysis. Although lifelog information can provide a customized and personalized setting for each user, it has some limitations: 1) it has to be compiled for each user separately, 2) it may have scalability problems as large-scale implementation of such infrastructure may not be practical, 3) there might be privacy issues related to gathering this kind of information, as users may not feel comfortable with being monitored this closely. Therefore, a more generic and yet personalized system is needed to propose personalized recommendations that respect user privacy and do not require complex infrastructure for its implementation. 

In this paper, we study how we can incorporate historical usage logs to build a data-driven recommender system for household lighting systems. Our aim is to increase the utility of using the enterprise lighting system as well as introducing advanced smart lighting features that can learn from household users' past behavior and incorporate user preferences and  psychology in the recommender system.We investigate user perceptions implicitly by leveraging past historical data and we make recommendations accordingly. In this manner, instead of proposing a control system which takes the control from the users, our system applies an implicit understanding of users so they can take charge of their environment via increased control over the lighting system. Recommended preferences and increased control over the lighting system makes users feel valued \cite{a2_gap}. As a result, the utility of using the system would increase. In addition, use of our system’s routine recommendations would probably result in less electricity usage. In this manner our proposed solution replaces prior lighting control systems and introduces a recommendation system that enables users with more control options.

\section{Methodology}
\label{S:2}
\subsection{Data abstraction}
The mathematical abstraction for describing the data is presented in this section.
The data set is based on actions (e.g., turn on/off, color change, rule setting,etc) and each observation defines an order that targets a light bulb. Light orders can come from various sources such as app, button and switches. All the orders go through a device called a hub which is responsible for the light-user interactions and for saving the communications between light bulb and the hub as log entries. Important features of each order considered in this analysis include timestamp associated with each order, color dimension features (saturation, brightness, color coordinates (x,y) , color temperature, and color mode), light id, group id that defines room type, and hub id which is the unique identification of a household, city and country. Overall this results in a 5th order tensor 
\begin{equation}
\label{x}
    X=\{x^{t_{1},d,r,b,s},x^{t_{2},d,r,b,s}, \ldots, x^{t_{N},d,r,b,s}\}  
\end{equation}
Where $x $ is a Boolean variable (on/off), $t_i$ denotes the $ith$ time instance in which $T=\{\,t_i \mid i \in N\,\}$,  $ s\in \{S\} $ denotes scenes  ,  $ d\in \{D\} $ denotes days ,  $ r\in \{room 1 , room 2\} $ denotes room type , $ b\in \{B\} $ denotes households.

\subsection{Routine recommender}
\begin{figure}[H]%
    \centering
    \subfloat[Unsupervised data driven threshold cut point.]{{\includegraphics[scale=0.55]{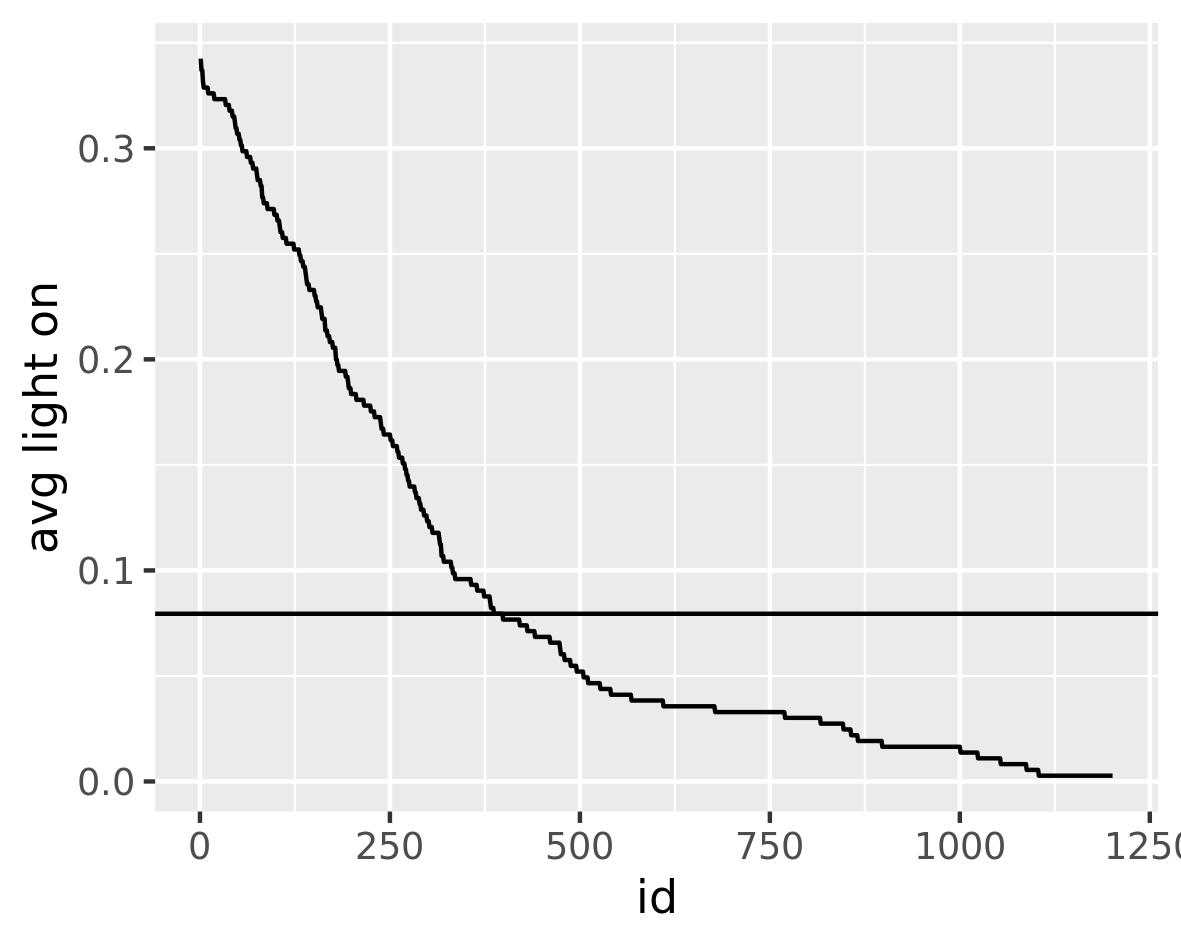} }}%
    \qquad
    \subfloat[random household 1 ]{{\includegraphics[scale=0.55]{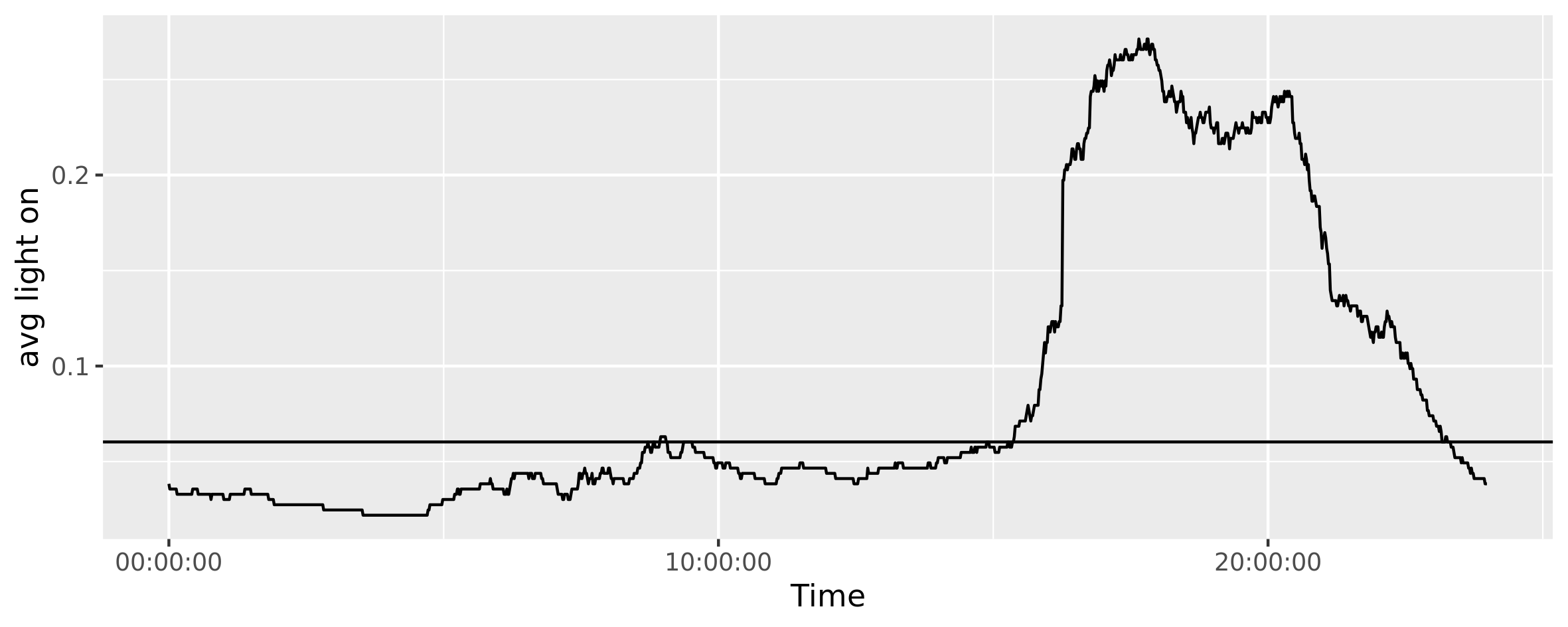} }}%
    \qquad
    \subfloat[random household 2]{{\includegraphics[scale=0.55]{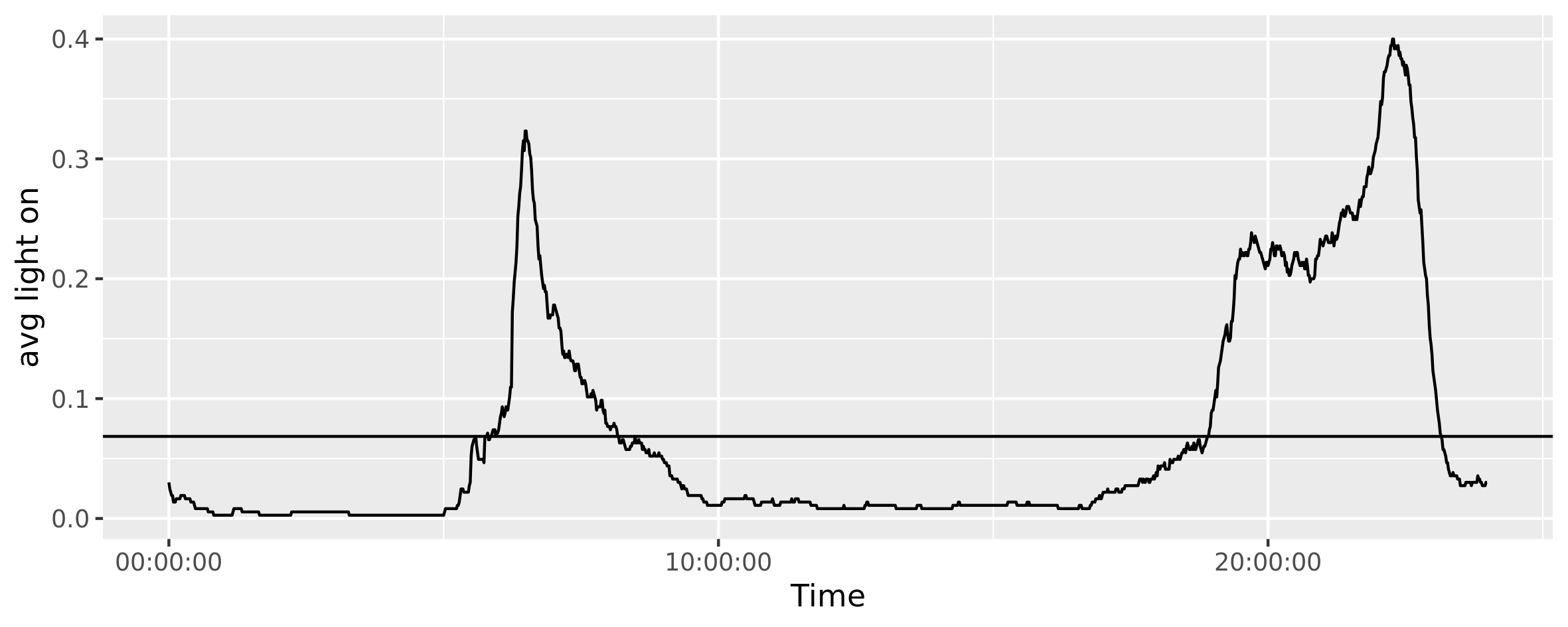} }}%
    \caption{Usage distribution and cutoff point}%
    \label{fig:schedule}%
\end{figure}

Our framework is separated into two major sub-systems, namely routine and scene recommenders. The auto routine recommender is based on frequency of light usage in each room type. We utilize an unsupervised learning method to select highly used periods of time. 
Frequency of use in each timestamp is defined using equation (\ref{2}) and is shown in figure \ref{fig:schedule}. Using the elbow cutoff method, as described in previous studies \cite{poddar2020massively,atousa:online} , shown in figure \ref{fig:schedule} sub figure (a) we define the margin in which the lights are highly used throughout the period. The horizontal line in figure \ref{fig:schedule} defines the margin identified by the elbow cutoff point, and the timestamps that have frequency above the threshold point are the desired  period of time based on usage. For instance, the recommended routine for this particular user in this room type occurs in the 7-10 PM time slot.

\begin{equation}
\ avg \ light \ on\ frequency=\frac{\sum\limits_{d=1}^{D}on }{D} , \forall t_i \in N 
\label{2}
\end{equation}

\subsection{Scene recommender}
\begin{figure}[H]
  \centering
    \includegraphics[scale=0.7]{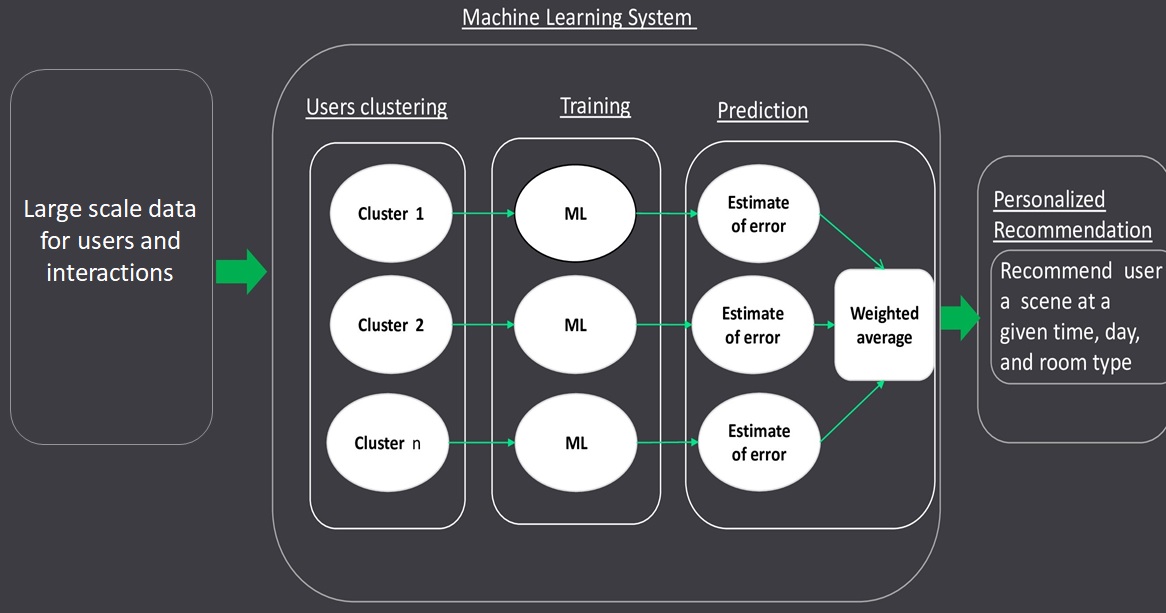}
    \caption{Proposed methodology for scene prediction}%
    \label{fig:method}%
\end{figure}

The scene recommendation methodology is shown in figure \ref{fig:method}. In the scene recommender system we cluster similar users and identify the most probable used scene in each hour based on similar users. Our sample consists of users located in four different countries across the world. We consider two highly used room types. We first cluster the users based on their usage and geographical location. Later on, using those clusters we train our machine learning model to predict the scene usage within each group of similar users. Finally we report the overall prediction accuracy via a weighted average over all clusters. We use multiple machine learning algorithms to explore the method before and after the clustering described above, and we present detailed descriptions of the methods in each section below.

\subsection{Data description and user clustering}
\label{section: data}
Our sample includes a total of 578 households located in United States (US) and abroad $b \in B\ $. This analysis considers two majorly used room types  as $r=\{\,r \mid r \in (room 1, room2)\,\}$ . The analysis is at the room level and therefore, all the scheduling, color recommendation and prediction are done at room level for each room type. As a result, each household can have a maximum of two room types in our analysis.
Our period of study was the 2019 calendar year, and the analysis is based on 8 predefined color environments defined as "scene" in this analysis $s \in S\ $.

\begin{figure}[H]
  \centering
    \includegraphics[scale=1]{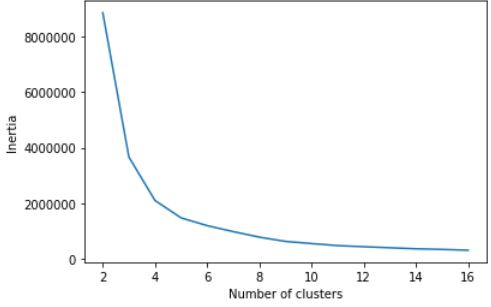}
    \caption{Optimum value of number of cluster.}%
    \label{fig:interia}%
\end{figure}

Clustering of similar users is based on usage characteristics and geographical location for each room type. We consider (0.15-0.85) quantiles values of "avg turn on frequency" defined in equation number \ref{2}, a vector of [1440 * 1 ] dimension (for 24 hour* 60 minutes) and one-hot encoding of country and room types as features for clustering. The number of clusters is defined using the elbow method over inertia values. With a result of three clusters corresponding to elbow point in Figure \ref{fig:interia}.

\begin{figure}[H]
  \centering
    \includegraphics[scale=0.6]{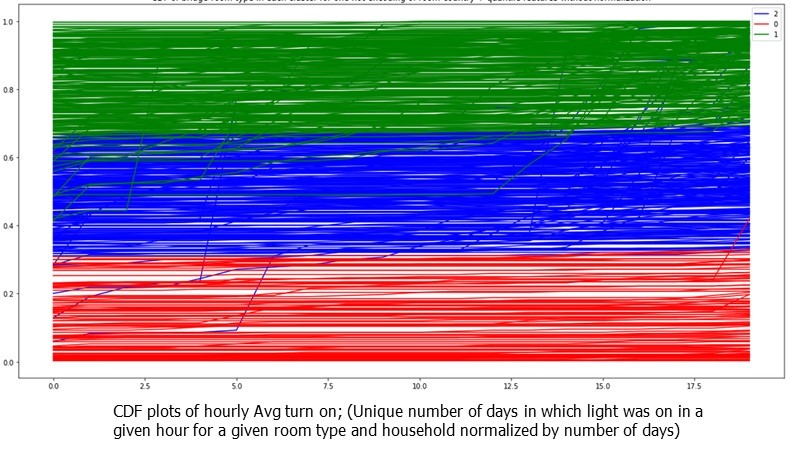}
    \caption{CDF plots for each household-room type colored by cluster number.}%
    \label{fig:cdf}%
\end{figure}

Later, in order to show the effectiveness of the clustering method, The distribution of usages in each household and room type is plotted. The CDF plots in figure \ref{fig:cdf} show the effectiveness of the method, as CDF lines are close to each other within each cluster and are separated from the other clusters. This represents the separation in distribution for each cluster.

\subsection{Features}
\label{section: features}
\begin{figure}[H]
  \centering
    \includegraphics[scale=0.3]{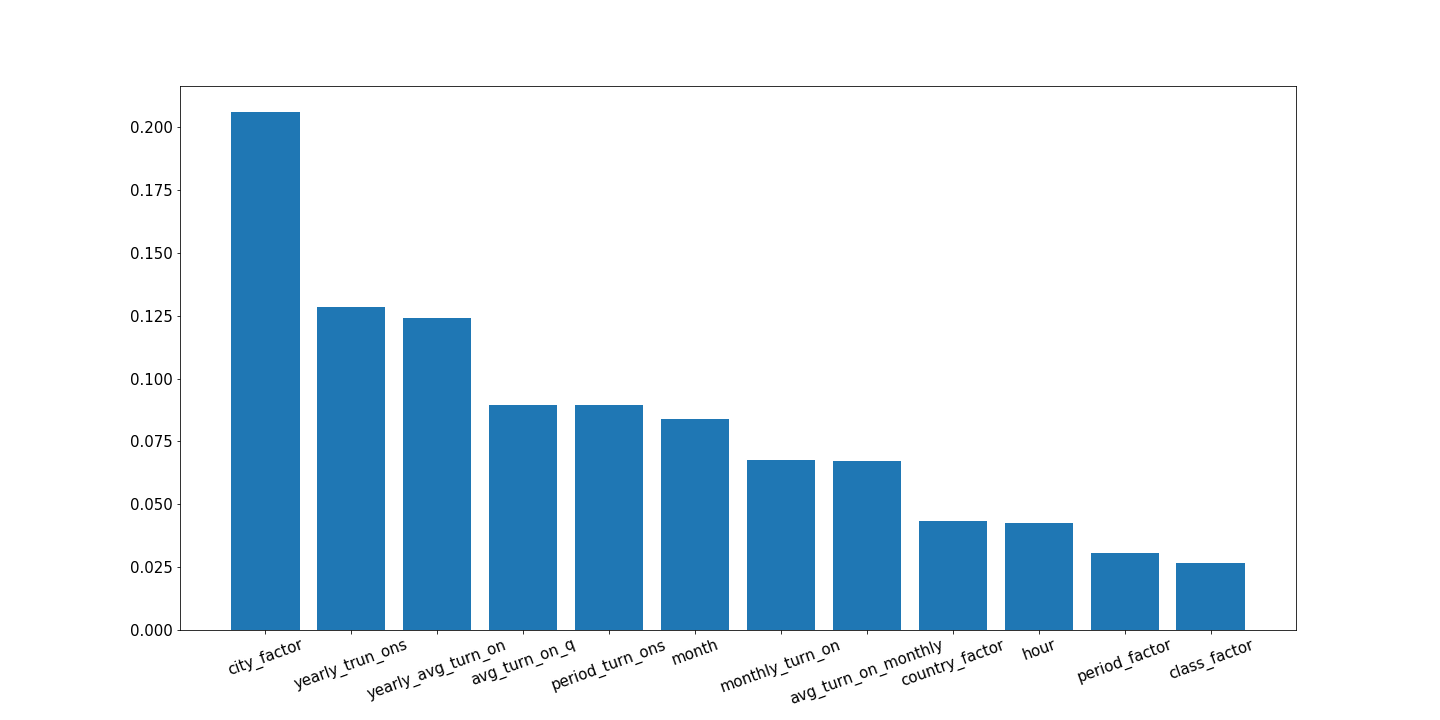}
    \caption{Feature importance}%
    \label{fig:feature imprtance}%
\end{figure}

Considering the methodology presented in figure \ref{fig:method}, a numerical description of the features are presented in table \ref{table: Numerical description of features}. The importance of each feature is presented in figure \ref{fig:feature imprtance}. The description of features used in both figure \ref{fig:feature imprtance} and table \ref{table: Numerical description of features} are as follows:

1- Monthly turn on : the unique number of days in which light was on in a given hour for a given room type and a particular household. 

2- Avg turn on : normalizes monthly turn on by dividing that number to the number of days available in the period of one month.

3- Quarterly turn on: the unique number of days in which light was on in a given quarter in a given hour for a given room type and particular household.

4- Avg turn on quarterly: normalized quarterly turn on divided by the total number of days available in a quarter. 

5- Yearly turn on: the unique number of days in which light was on at a given hour for a given room type and particular household.

6- Yearly avg turn on: normalized yearly turn on divided by the total number of days in the year. 

We also consider temporal characteristics such as " month" , "hour" and "period factor". Geographical features are presented as "city factor" and "country factor". Finally "class factor" defines the room type in this analysis.

\begin{table}[H]
\caption{Numerical description of features}
\label{table: Numerical description of features}
  \setlength\tabcolsep{6pt} 
  \footnotesize
    \begin{tabularx}{\textwidth}{c*{8}{Y}}
      \hline
 & month & hour & monthly \ turn on & avg turn on  monthly & quarterly\ turn on & avg turn on\ quarterly & yearly \ turn on & yearly avg \ turn on\\ \hline
mean & 5.83  & 11.5 & 22.1              & 0.74                   & 61.52             & 0.68             & 219.26            & 0.68                \\
std  & 3.09  & 6.92 & 8.42              & 0.28                   & 23.64             & 0.26             & 71.19             & 0.22                \\
min  & 1     & 0    & 1                 & 0.03                   & 1             & 0.01             & 1              & 0.00                \\
25\% & 3     & 6    & 16                & 0.53                   & 44             & 0.49             & 170            & 0.52                \\
50\% & 6     & 12   & 25                & 0.83                   & 65             & 0.72             & 228            & 0.70                \\
75\% & 8     & 18   & 30                & 1                   & 84            & 0.93             & 283           & 0.87                \\
max  & 11    & 23   & 31                & 1.03                   & 92             & 1.02             & 324           & 1                \\ \hline
    \end{tabularx}
\end{table}

\subsection{Training and prediction}
We utilize random forest , KNN, and xgboost to compare the prediction performance across different classes. Random forest is a  tree-based meta estimator that fits an estimator on various sub samples and reduces the prediction variance and prevents overfitting by averaging over the models trained on the sub samples. Sub sample size is controled with the bootstrap option. XGBoost is an ensemble method of gradient boosted trees and it works by combining weak predictive tree models and learning from them. KNN implements k nearest neighbors and voting among neighbors. In this study, the above mentioned methods' parameters were optimized based on grid search and are as follow: number of tree, max depth of trees, and number of neighbours. We split the data to train-test split and use independent test and cross validation for evaluating the prediction.

\subsection{Evaluation}

Here we analyze the performance of different algorithms described in methodology, before and after the clustering. Using data described in section \ref{section: data} and features described in section \ref{section: features} we aim to provide a personalized scene recommendation. To do so, we first compare different algorithms and their prediction performance for each of our classes (colors). We used precision, recall, F-score, accuracy and balanced accuracy in our framework for evaluation purposes in Equations (\ref{recall}-\ref{balanced ac}) 

\begin{itemize}
\item (TP):True Positive
\item (TN):True Negative
\item (FP):False Positive
\item (FN):False Negative
\end{itemize}

\begin{equation}
Recall = \frac{TP}{TP+FN}
\label{recall}  
\end{equation}

\begin{equation}
Specificity = \frac{TN}{TN+FP}
\label{specificity}  
\end{equation}

\begin{equation}
Precision = \frac{TP}{TP+FP}
\label{precision}  
\end{equation}

\begin{equation}
Accuracy=\frac{TP+TN}{TP+FP+TP+TN}
\label{accuracy}  
\end{equation}

\begin{equation}
F1-score=\frac{2 * (Precision * Recall)}{Precision + Recall}
\label{f1}  
\end{equation} 

\begin{equation}
Balanced \ Accuracy =\frac{Recall+Specificity}{2}
\label{balanced ac}  
\end{equation} 

\section{Results and discussion}
\subsection{General recommendation}

In this study we analyzed the performance of each category (color) in our multi category classification problem. We then provided a summary result for each method both in terms of accuracy and balanced accuracy. Later, by selecting the algorithm, we compared the performance of the algorithm for the two phases of before and after clustering. Moreover, we analyzed the accuracy and balanced accuracy within each separate cluster. 
The results of our model are described in this section in three stages: First, prior to clustering, we set a benchmark by analyzing different algorithms and their performance on our set up. Second, we summarized the results of classifier estimates into two single metrics of accuracy and balanced accuracy. Third, we trained our algorithm on each cluster separately and reported the results in terms of accuracy and balanced accuracy and aggregated them.
In the first stage, we randomly split the data into train and test (90 \% -10 \%) sets. Table \ref{Performance metrics} shows the results of different algorithms in terms of recall, precision and F-score metrics on our test set using each of the classification methods discussed in section 3.6.

\begin{table}[H]
    \centering
    \caption{Performance metrics in different algorithms (test set) }
\begin{tabular}{cccccccccc}

\hline
\multicolumn{1}{l}{} & \multicolumn{3}{c}{Precision} & \multicolumn{3}{c}{Recall} & \multicolumn{3}{c}{F-score} \\ \hline
                class/method     & RF      & KNN     & XGBoost   & RF     & KNN    & XGBoost  & RF     & KNN    & XGBoost  \\ \hline
0                    & 0.97    & 0.94    & 0.92      & 0.97   & 0.96   & 0.95     & 0.97   & 0.95   & 0.93     \\
1                    & 0.97    & 0.94    & 0.93      & 0.91   & 0.94   & 0.93     & 0.94   & 0.94   & 0.93     \\
2                    & 0.94    & 0.90    & 0.94      & 0.88   & 0.87   & 0.88     & 0.91   & 0.89   & 0.91     \\
3                    & 0.96    & 0.93    & 0.96      & 0.99   & 0.91   & 0.91     & 0.97   & 0.92   & 0.93     \\
4                    & 0.96    & 0.87    & 0.95      & 0.94   & 0.81   & 0.94     & 0.95   & 0.84   & 0.95     \\
5                    & 1.00    & 0.92    & 0.96      & 1.00   & 0.81   & 1.00     & 1.00   & 0.86   & 0.98     \\
6                    & 0.97    & 1.00    & 0.97      & 0.89   & 1.00   & 0.91     & 0.93   & 1.00   & 0.94     \\
7                    & 1.00    & 0.84    & 1.00      & 1.00   & 0.79   & 1.00     & 1.00   & 0.82   & 1.00     \\
8                    & 1.00    & 0.71    & 1.00      & 0.87   & 0.71   & 1.00     & 0.93   & 0.71   & 1.00     \\ \hline
\label{Performance metrics}
\end{tabular}
\end{table}
The scene prediction problem is a multi category classification problem and, therefore, classifier performance should be judged based on each separate category. The category prediction results presented in table \ref{Performance metrics} confirmed that there was no crucial imbalance in the prediction. Moreover, performance was relatively robust with respect to each category. 

In the second stage, classifier estimate results provided for each category in table \ref{Performance metrics} were summarized into two single metrics (accuracy and balanced accuracy) as shown in table \ref{after clustering summary} .

\begin{table}[H]
\centering
\caption{Test accuracy before clustering (test set)}
\begin{tabular}{ccc}
\hline
Method  & Accuracy & Balanced accuracy \\ \hline
RF      & 0.965    & 0.939             \\
KNN     & 0.9312   & 0.86              \\
XGboost & 0.928    & 0.946             \\ \hline
\label{after clustering summary}
\end{tabular}
\end{table}

In the third stage, using features and method described in section \ref{section: data}, we segmented our sample into three clusters. Considering F1-score and accuracy we used random forest as our classifier. Within each cluster, we split our data into train-test (90\% - 10\%) sets and  reported the results in terms of accuracy, balanced accuracy and weighted average accuracy over the clusters populations in table \ref{weighted}.  As shown in table \ref{weighted}, the overall weighted average has a meaningful increase both in terms of accuracy and balanced accuracy. This shows that training within similar users enhances the recommendation system by 0.72 percent in terms of accuracy.

\begin{table}[H]
\centering
\caption{ Weighted test accuracy after clustering}
\begin{tabular}{cccc}
\hline
Cluster number           & Balanced accuracy & Accuracy & Cluster Population \\ \hline
0                        & 0.98              & 0.987    & 133                \\
1                        & 0.93              & 0.972    & 259                \\
2                        & 0.94              & 0.965    & 263   
\\ \hline
Overall weighted average & 0.944             & 0.972    & 655                \\ \hline
\label{weighted}
\end{tabular}
\end{table}

\subsection{Cold start}
The lack of prior knowledge on user preferences causes recommendation systems to face a problem known as cold start. Here we analyze cold start in our scene prediction problem and we report our efforts to cope with this issue. We show that clustering can increase the prediction accuracy in a cold start context. Figure \ref{random sampling} shows the effect of sampling size on model performance when facing cold start. The train and test sets include different households and therefore, it reflects the condition of no prior knowledge on specific household preferences. 

We investigated three test sets with different sample sizes as follows: In scenario 1, we reserved 10\% of households for the test set and trained on 90\% . In scenario 2, we tested on 25\% of households and trained on 75\%. 
In scenario 3, we reserved 40\% of households for the test set and trained on 60\% . We ran 20 randomly sampled iterations using this set up for each scenario and reported cross validation averaged accuracy on train and test sets. In addition, we obtained independent test prediction accuracy as shown in figure \ref{random sampling}.

The average accuracy over the independent test prediction and test cross validation increases by increasing the sample size. Moreover, the increase in training sample size results in lower deviation in test accuracies for both cross validation and independent test prediction. Cross validation shows better performance in terms of average test set accuracy in the first scenario where the test size is smaller. As with smaller datasets it is better to do cross validation and train on entire dataset. On the other hand, when running independent testing, the average accuracy increases by increasing the training sample size. As such, by increasing the test sample size to 40 \% we were able to achieve a more robust model. Overall, the comparison of cross validation accuracy in train sets in the three scenarios suggests that having more data would result in better performance in terms of accuracy and with smaller dataset deviation of prediction accuracy increases.

\begin{figure}[H]
\label{random sampling}
  \centering
    \includegraphics[scale=0.8]{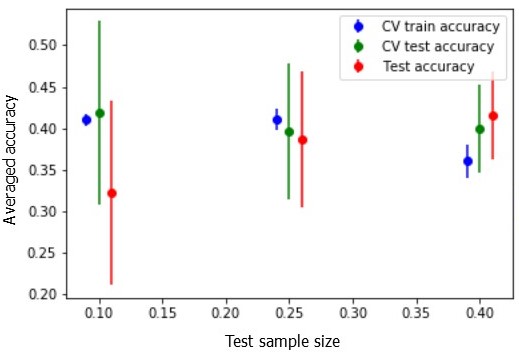}
    \caption{Random sampling on household separation result.}%
    \label{random sampling}%
\end{figure}

 Table \ref{'table: clustering cols start'} shows effect of clustering on how well we are able to predict the preferences in scene recommendations for the cold start problem. As shown after clustering, accuracy has a meaningful increase and it is beyond the deviation in the classifier. It is notable that accuracy in clustering method is reported by a weighted average over the population of each cluster.

\begin{table}[h]

\centering
\begin{tabular}{cccc}
\hline
 Clustering    &    Metric           & Mean & Standard deviation \\ \hline
Before & CV test             & 0.41 & 0.03               \\
After & CV weighted average & 0.44 & 0.02               \\ \hline
\end{tabular}

\caption{Cold start problem and clustering}
\label{'table: clustering cols start'}
\end{table}
 It should be noted that lighting needs and preferences in each room type and household are affected by many personal and environmental factors. The personal factors include mood, general activity, character, and living style while the environmental factors include design of building for each room type, apartment level, window size and direction, and level of daylight for each room. When there is no prior knowledge of users' environmental and personality factors, a time-location based prediction of users' desired light color can be challenging.
Thus a more accurate prediction would requires access to personal information that could interfere with user privacy.
 
 

\section{Conclusion and future work}
Every individual's preference and tolerance 
in terms of lighting and color needs is different \cite{tregenza1974consistency} and, even for a particular user, different tasks may require different lighting settings.
Previous research has revealed that working under desired lighting can increase satisfaction levels and productivity \cite{juslen2007influence}. Therefore, providing personalized lighting recommendations could increase user satisfaction and engagement with the system. 
 
In this paper we aimed to include recommendation as a smart add-in feature in the organization's product in order to increase customer utility, loyalty and satisfaction when interacting with lighting systems. In this study, as opposed to many studies found in the literature that developed light control systems for office locations, we proposed a recommendation system for households needs. This problem consisted of two subproblems: light schedule and color, which we dealt with as routine and scene recommendations. We proposed separate frameworks for each subproblem as follows: For routine recomendation, we  used an unsupervised clustering method based on frequency of light usage in each minute of the period of analysis. We considered the scene recomendation problem as a multi-category classification problem and we leveraged historical data from households located in different geographical locations around the world for a period of one year to develop personalized recommendations. We utilized clustering to enhance our system's prediction performance and we analyzed the performance of our classifier on two different settings(with and without prior knowledge of household preferences). Results revealed that training on similar users enhanced our classifier's prediction performance on  generic and cold start recommendations.

Mood and emotion affect how humans perceive and interact with lighting systems. Therefore,
identifying human mood, behavior, psychology, and emotions at each time using other sources and associating those pieces of information with a recommendation system is a promising line of future research work.







\bibliographystyle{elsarticle-num-names}
\bibliography{sample.bib}







\end{document}